\pdfoutput=1

\documentclass{article}

\usepackage{arxiv}  % <--- Include the .sty file if you actually need this.

\usepackage[utf8]{inputenc}    % Ensure UTF-8 encoding
\usepackage{fourier}           % Fourier fonts (optional, can remove)

\usepackage{graphicx}          % For including graphics
\usepackage{booktabs}          % Better table rules
\usepackage{multirow}          % Multi-row tables
\usepackage{colortbl}          % For colored cells in tables
\usepackage[table,dvipsnames]{xcolor} % Consolidated xcolor options here
\usepackage{comment}           % Commenting out large sections
\usepackage{wrapfig}           % Wrap text around figures

\usepackage{fontawesome}       % May need to include .sty if not default
\usepackage[many]{tcolorbox}   % May need local .sty if arXiv doesn't have latest
\usepackage{amsmath, amssymb, mathtools, amsthm}
\usepackage{array}             % Advanced array & tabular features

\usepackage[square,numbers]{natbib}  % Works if you provide a .bbl file

\usepackage{nicefrac}          % Compact symbols for 1/2, etc.
\usepackage{soul}              % Underlining, etc.
\usepackage{doi}               % Makes DOIs clickable

\usepackage{caption}
\usepackage{subcaption}

\usepackage{enumitem}

\usepackage{hyperref}

\title{Artificial Intelligence in Spectroscopy: Advancing Chemistry from Prediction to
Generation and Beyond}

\definecolor{CalGoldHex}{HTML}{FDB515}
\definecolor{darkblue}{HTML}{185f8c}
\definecolor{gold}{HTML}{95711e}

\newcommand\blfootnote[1]{
  \begingroup
\renewcommand\thefootnote{}\footnote{#1}%
  \addtocounter{footnote}{-1}%
  \endgroup
}

\newcommand*{\affmark}[1][*]{\textsuperscript{\textnormal{#1}}}

\author{
\textbf{Kehan Guo}\affmark[1], \textbf{Yili Shen}\affmark[1], 
\textbf{Gisela Abigail Gonzalez-Montiel}\affmark[2],
\textbf{Yue Huang}\affmark[1], 
\textbf{Yujun Zhou}\affmark[1], \\ \textbf{Mihir Surve}\affmark[2], \textbf{Zhichun Guo}\affmark[3], \textbf{Prayel Das}\affmark[4], \textbf{Nitesh V Chawla}\affmark[1], 
\textbf{Olaf Wiest}\affmark[2], \textbf{Xiangliang Zhang}\affmark[1]\\
~\\
\affmark[1]Department of Computer Science and Engineering, University of Notre Dame~~~\\\affmark[2]Department of Chemistry and Biochemistry, University of Notre Dame~~~\\
\affmark[3]Institute for Protein Design, University of Washington~~~\\
\affmark[4]Trusted AI Department of IBM Thomas J Watson Research Center, IBM~~~
\\
}

\definecolor{lightblue}{RGB}{173, 216, 230}

\renewcommand{\shorttitle}{Artificial Intelligence in Spectroscopy: Advancing Chemistry from Prediction to
Generation and Beyond}

\newcolumntype{C}[1]{>{\centering\arraybackslash}m{#1}}
\newcolumntype{L}[1]{>{\raggedright\arraybackslash}m{#1}}

\definecolor{deepred}{rgb}{0.631,0.102,0.102}
\definecolor{amethyst}{rgb}{0.6, 0.4, 0.8}
\definecolor{darkgreen}{rgb}{0.3,0.7,0.3}
\definecolor{salmon}{RGB}{241, 150, 141}

\hypersetup{
    colorlinks=true,  % Change citation color to blue
    citecolor = gold,
    linkcolor=blue,
    urlcolor=blue 
}

\begin{document}
\maketitle
\blfootnote{Correspondence to: Xiangliang Zhang~(\url{xzhang33@nd.edu}).}

\begin{abstract}
The rapid advent of machine learning (ML) and artificial intelligence (AI) has catalyzed major transformations in chemistry, yet the application of these methods to spectroscopic and spectrometric data--termed \textit{Spectroscopy Machine Learning (SpectraML)}--remains relatively underexplored. Modern spectroscopic techniques (MS, NMR, IR, Raman, UV-Vis) generate an ever-growing volume of high-dimensional data, creating a pressing need for automated and intelligent analysis beyond traditional expert-based workflows. In this survey, we provide a unified review of SpectraML, systematically examining state-of-the-art approaches for both \textit{forward} tasks (molecule-to-spectrum prediction) and \textit{inverse} tasks (spectrum-to-molecule inference). We trace the historical evolution of ML in spectroscopy--from early pattern recognition to the latest foundation models capable of advanced reasoning--and offer a taxonomy of representative neural architectures, including graph-based and transformer-based methods. Addressing key challenges such as data quality, multimodal integration, and computational scalability, we highlight emerging directions like synthetic data generation, large-scale pretraining, and few- or zero-shot learning. To foster reproducible research. We also release an open-source repository listing recent papers and their corresponding curated datasets\footnote{\url{https://github.com/MINE-Lab-ND/SpectrumML_Survey_Papers}}. Our survey serves as a roadmap for researchers, guiding advancements at the intersection of spectroscopy and AI.
\end{abstract}

\section{Introduction}
\label{sec:intro}

The rapid advancements in Artificial Intelligence (AI) and Machine Learning (ML) are reshaping scientific disciplines of chemistry, by streamlining tasks such as molecular property prediction~\citep{guo2021few} and reaction modeling~\citep{coley2019graph}. Despite these breakthroughs, the application of ML to spectroscopy--hereafter referred to as \textbf{Spectroscopy Machine Learning (SpectraML)}~\citep{elias2004intensity,ralbovsky2020towards}--remains relatively underexplored. Spectroscopic and spectrometric techniques, which provide high-sensitivity insights into molecular structure, dynamics, and properties, are now generating large volumes of data due to advances in high-throughput experiments and automated acquisition. Consequently, traditional manual analysis methods, reliant on expert interpretation and reference libraries~\citep{alberts2024unraveling,zhu2023rapid}, are increasingly inadequate for handling the scale and complexity of modern spectral datasets.

\begin{figure}[t!]
    \centering
\includegraphics[width=0.9\linewidth]{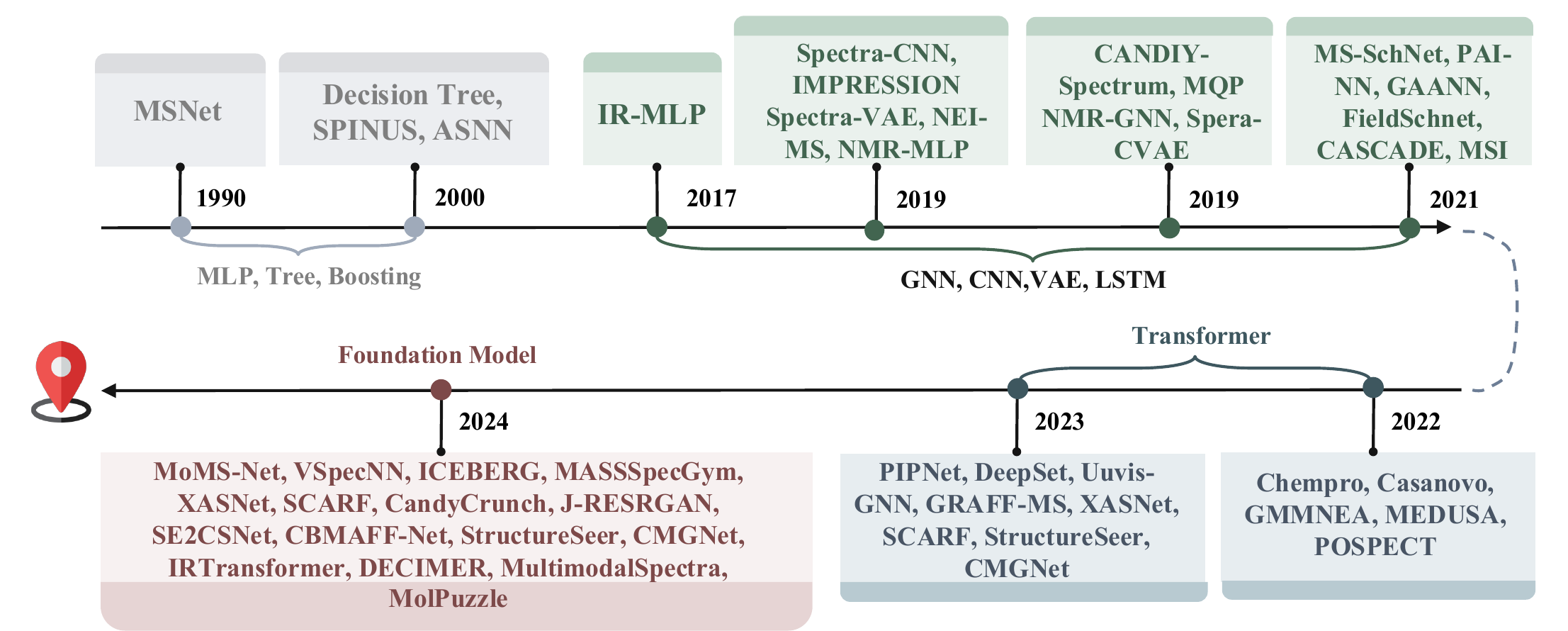}
    \caption{Timeline of ML progression and its application to spectroscopic studies. Left: Molecule to Spectrum, Right: Spectrum to Molecule} 
    \label{fig:timeline}
    \vspace{-0.15in}
\end{figure}

The growing interest in this field is reflected in the increasing number of research papers that expand the scope of tasks addressed by spectrum-based ML models~\citep{gabrielprospect,alberts2024unraveling,guo2024can,bushuiev2024massspecgym}. However, existing overviews often focus on a single modality (e.g., UV alone or MS)~\citep{beck2024recent} or lack a clear framework for distinguishing forward (molecule-to-spectrum) from inverse (spectrum-to-molecule) tasks~\citep{sridharan2022modern}. By contrast, our survey unifies five major spectroscopic techniques--MS, NMR, IR, Raman, and UV-Vis--within a single methodological framework. Moreover, we highlight the rapid progression of spectroscopic analysis driven by ML advances %emphasize emerging directions 
in generative modeling, few- or zero-shot learning, and large-scale pretraining, and we provide an open-source repository of datasets and code. By bridging computational and experimental viewpoints, our work fills a key gap in the literature and highlights new avenues for interdisciplinary collaboration in SpectraML.

The rapid advancements in ML and AI have been transforming workflow automation in spectral analysis, as illustrated by the timeline in Fig. \ref{fig:timeline}. Deep learning models, such as convolutional neural networks (CNNs)~\citep{o2015introduction} and recurrent neural networks (RNNs)~\citep{schuster1997bidirectional}, have proven effective in tasks like peak detection, deconvolution, and reaction monitoring. Additionally, transfer learning and pre-trained models~\citep{guo2024can} enable these algorithms to generalize across diverse spectra, thereby reducing the need for extensive retraining. Emerging foundation models~\citep{bommasani2021opportunities} further extend the capabilities of SpectraML by offering advanced reasoning and planning for complex tasks such as molecular structure elucidation and reaction pathway prediction~\citep{guo2024can}. As AI techniques continue to evolve, there is a critical need for a structured discussion on positioning the different capabilities of AI models across various spectroscopy tasks, as well as underscoring key challenges, limitations, and future directions. %Yet, previous efforts in this domain have often been fragmented and lack a comprehensive, unified review that integrates these advancements.

This survey addresses these needs with the following contributions:
%by providing a unified review of recent SpectraML studies. Our contributions are threefold:
\begin{enumerate}[nolistsep, leftmargin=*]
    \item We offer a comprehensive overview of current SpectraML techniques across five major spectroscopic modalities--MS, NMR, IR, Raman, and UV-Vis--highlighting both methodological innovations and practical applications. Unlike existing surveys that focus on a single modality or overlook the distinction between forward (molecule-to-spectrum) and inverse (spectrum-to-molecule) tasks~\citep{beck2024recent,sridharan2022modern}, our work provides a unified perspective and frames these tasks within AI's problem-solving role.
    \item We present a \textbf{unified roadmap} that traces the evolution of ML in spectroscopy,  from early pattern recognition and predictive analytics to advanced generative and reasoning frameworks, thus situating current progress within a broader historical context. It helps researchers understand how foundational techniques have shaped modern approaches and guides them in innovating future methodologies in SpectraML.
    \item We identify key challenges (e.g., data quality, multimodal integration, and computational scalability) and emerging opportunities (e.g., foundation models, synthetic data generation, few- or zero-shot learning, and large-scale pretraining) in SpectraML. To facilitate further research, we provide and will maintain, an open-source \href{https://github.com/YujunZhou/SpectrumML_Survey_Papers}{GitHub repository} containing datasets and code. This work thus serves as a valuable resource for researchers and practitioners in this interdisciplinary field. %By bridging computational and experimental viewpoints, our survey paves the way for interdisciplinary collaboration and future advancements in SpectraML.
\end{enumerate}

The remainder of this paper is organized as follows. Section~\ref{sec:background} introduces spectral data representations and the definition of fundamental \textbf{Forward and Inverse Problems} in spectral analysis. Section~\ref{sec:methodology} categorizes and summarize SpectraML approaches in solving forward and inverse problems. %, from viewpoints of involved molecular and spectral representations, covering various ML methods such as \textbf{graph-based models}, and \textbf{transformer-based architectures}. 
In Section~\ref{sec:discussion}, we discuss major challenges %in data quality, multimodal integration, and computational scalability, 
and highlight emerging directions such as \textbf{foundation models}, and \textbf{synthetic data generation}. %, and \textbf{hybrid modeling approaches}. 
%Finally, 
Section~\ref{sec:conclusion} concludes the work.

\section{Background}
\label{sec:background}

\subsection{Applications of Spectroscopy in Chemistry }

Spectroscopy, the study of the interaction between matter and electromagnetic radiation, produces data that resembles \textbf{audio signals} in its representation--peaks, shifts, and patterns that encode molecular information \citep{elias2004intensity}. Spectrometry, on the other hand, focuses on measuring chemical interaction to gain insight into molecular structures and properties \citep{ralbovsky2020towards}. Common spectroscopic techniques include mass spectrometry (MS), infrared (IR), Raman, ultraviolet/visible (UV-Vis), and nuclear magnetic resonance (NMR). Each of these techniques is akin to a "lens" providing a different perspective of the molecular world, and when combined, they reveal a fuller picture of molecular structures.
%Each technique offers unique capabilities:
\begin{itemize}[nolistsep, leftmargin=*]
    \item \textbf{Mass Spectrometry (MS)} allows for the determination of the molecular mass and  formula of a compound, as well as some of its structural features by identifying the fragments produced when the molecule breaks apart. 
    \item \textbf{Infrared (IR)} and \textbf{Raman} spectra data allow the identification of the types of functional groups in a compound.
    \item \textbf{UV-Vis}  spectra data provides information about compounds that have conjugated double bonds.
    \item \textbf{NMR}  spectra data provide information about atomic nuclei (e.g., the carbon-hydrogen framework of a compound).  Advanced  techniques,  2D and 3D-NMR, further enable the characterization of complex molecules such as natural products, proteins, and nucleic acids.
\end{itemize}

The obtained spectra data are widely used across chemistry, biology, and related fields, akin to a "molecular microscope" that enables researchers to explore the unseen. These spectral data are often presented as plots or graphs that visually represent the relationship between intensity and a specific variable, such as wavelength, wavenumber, or mass-to-charge ratio (\textit{m/z}), as demonstrated on the left part of Figure \ref{fig:task}. The graphic format details of these spectral data can be found in the Appendix \ref{sec:Spectroscopy-data}.  Studies involving these data are generally divided into two main categories:
\begin{itemize}[nolistsep, leftmargin=*]
    \item \textbf{\textit{Forward  Problem}}: \textit{predicting a spectrum based on molecular structure information}. While spectroscopy devices can generate spectra from molecular samples, solving the \textbf{Forward  Problem} (structure-to-spectrum problem) using AI  models is highly valuable and offers several key advantages. First, it reduces the need for costly and time-consuming experimental measurements by enabling rapid spectral predictions. Second, it enhances the understanding of fundamental relationships between molecular structures and their spectral signatures. Such structure-to-spectrum correlation is crucial for scientists to know what molecule(s) are present for drug discovery, biomarker research, natural product synthesis, and other research areas \citep{mandelare2018coculture}. Lastly, it expands applications beyond experimental limits.  Some molecules are difficult to analyze using standard spectroscopy due to low concentrations, unstable intermediates, or extreme environmental conditions. AI solutions enables insights into such challenging cases where direct measurement is impractical.  %Here, the major challenge is a two-part pipeline. The first part lies in the correct translation and interpretation of chemical properties from 2D and 3D molecular structures. The second part lies in taking the extrapolated chemical information from the first step and correctly generating the corresponding spectral data of that molecule. This is where AI/ML can rapidly multi-task, generate several plausible outcomes, and accurately predict spectra figures that have chemical meaning. 
    \item \textit{\textbf{Inverse (Backward) Problem}: deducing the molecular structure based on experimentally obtained spectra,} also known as molecule elucidation,  is a crucial task in life sciences, chemical industries, and other fields \citep{sridharan2022modern,yao2023conditional}.
    Resolving this problem enables researchers to identify unknown compounds, verify chemical compositions, and gain deeper insights into molecular behavior, ultimately advancing scientific discovery and industrial applications. However, molecular elucidation remains a time-consuming and complex process that heavily relies on human expertise. Identifying spectrum-to-structure correlation is particularly challenging, requiring analysts to distinguish real peaks and accurately deduce their chemical meaning. Manual interpretation is labor-intensive, has limited scalability, and is also prone to misinterpretation due to overlapping signals, sample impurities, and isomerization issues. This is where AI   can play a transformative role, automating spectral interpretation, and accelerating the resolution of inverse problems.  
   % Here, the major challenge lies in interpreting spectral data--a time-consuming process prone to misinterpretation and characterization by errors arising from the sample (e.g., overlapping signals, purity or mixture, isomerization issues) or errors from the instrument (e.g., noise, calibration, data acquisition). However, assuming that spectra data are clear from these errors, it is still time-consuming to identify the real peaks and deduce the chemical meaning of those peaks into a molecular structure. This is where AI/ML can play a transformative role and accelerate spectral analyses.
\end{itemize}
Note that the above definition of the forward/inverse problem is in accordance with what is commonly referred to in the community \citep{lu2024deep}. However, the opposite definition exists in some contexts, e.g., in \citep{beck2024recent}, where the inverse problem focuses on predicting spectra, while the forward problem refers to molecular deduction from given spectra. This difference in terminology highlights the slightly varying perspectives across disciplines and underscores the need for clear definitions when discussing these concepts in the context of spectroscopy and ML applications.

\subsection{Roadmap of  SpectraML} %Machine Learning in Spectra Analysis}

% adding examples 

ML has revolutionized the way spectroscopic data is analyzed, offering new pathways to extract deeper insights, accelerate workflows, and uncover patterns beyond human capability. Historically, the use of computational techniques in spectroscopy was limited to basic pattern recognition and property prediction tasks ~\citep{elias2004intensity}. This changed with the advent of deep learning and advanced ML frameworks that have enabled transformative capabilities across the entire spectrum analysis pipeline. For instance, CNNs excel in tasks such as peak detection~\citep{o2015introduction} and deconvolution~\citep{hu2024accurate}, akin to identifying features in an image, while RNNs and transformers~\citep{schuster1997bidirectional} handle sequential spectral data, similar to interpreting audio signals, making them suitable for reaction monitoring and dynamic studies. %These advances have led to significant breakthroughs in both chemistry and biology. 
For example, CASCADE~\citep{guan2021real} accelerates the prediction of chemical shift predictions in NMR spectra by \textit{6000 times} comparing to the fastest DFT method, enabling real-time NMR chemical shift predictions from simple molecular representations.

As spectroscopic datasets have grown in size and complexity, ML has demonstrated exceptional scalability and adaptability. The shift from early predictive models to modern \textbf{generative} and \textbf{reasoning} frameworks, such as attention-based transformers and foundation models, has redefined the scope of spectral analysis. Generative models enable the simulation of spectra based on molecular structures ~\citep{goldman2023prefix}, addressing the \textbf{forward problem}, while reasoning-driven models tackle the \textbf{inverse problem}, predicting molecular structures with enhanced accuracy~\citep{alberts2024leveraging,alberts2023learning}. More discussion regarding these two types of problems is presented in the next section. These developments have brought unprecedented precision and speed to applications ranging from molecular characterization to reaction pathway prediction.
For example, IMPRESSION~\citep{gerrard2020impression} predicts NMR parameters with near-quantum chemical accuracy while accelerating computational time from days to seconds.

\begin{figure*}[t]
\centering
    \includegraphics[width=1\linewidth]{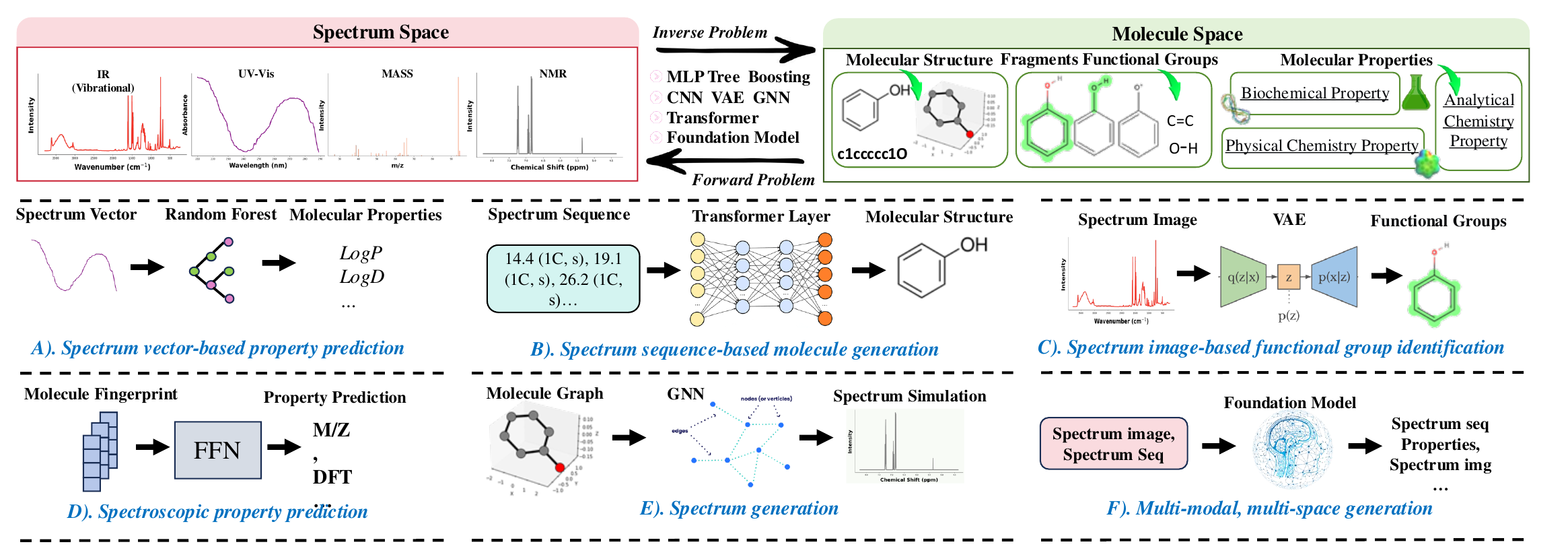} \vspace{-0.1in}
    \caption{(Top) Overview of SpectraML, translating between \textbf{Spectrum Space} and \textbf{Molecule Space}. %, focusing on inverse (spectrum to molecule) and forward (molecule to spectrum) problems. 
    (Middle and Bottom) Illustration of key tasks in SpectraML, including their \textbf{inputs}, \textbf{outputs}, and the \textbf{machine learning models} used for mapping them, such as Random Forest, Feed Forward   Networks (FFN), Variational Autoencoders (VAE), Transformers,  Graph Neural Networks (GNN), and Foundation Models.  } %\textbf{A)} Spectrum vector-based property prediction, \textbf{B)} Spectrum sequence-based molecule generation, \textbf{C)} Spectrum image-based functional group identification, \textbf{D)} Spectroscopic property prediction using fingerprints, \textbf{E)} Spectrum generation from molecular graphs, and \textbf{F)} Multi-modal, multi-space generation via foundation models. }
    \label{fig:task}
    \vspace{-0.1in}
\end{figure*}

\section{SpectraML Methodologies Summary}
%Machine Learning Methodologies for Spectral Analysis}
\label{sec:methodology}

In this section, we present a detailed discussion of the machine learning methodologies that address the twin challenges: the \textbf{forward (molecule-to-spectrum) and inverse (spectrum-to-molecule) problems}. Our discussion is organized around four core components. We begin by examining the data representations and preprocessing strategies that serve as the foundation for effective spectral modeling. We then focus on the forward problem of predicting spectral signatures from molecular structures, followed by a discussion of the inverse problem of inferring molecular structures from spectral data. Finally, we describe emerging unified frameworks and cross-modal integration approaches that promise to address both challenges simultaneously. %This integrated perspective not only highlights methodological innovations but also establishes direct links between these techniques and practical research problems.
%encountered in SpectraML (
A summary of the discussed work is presented in Table \ref{tab:paper_summarization}.

\subsection{Data Representations and Preprocessing}

The quality of spectral analysis is fundamentally determined by how both molecular and spectral data are represented and preprocessed. In SpectraML, spectral data may be expressed as vectors, sequences, or images, while molecular structures are encoded using vector-based descriptors, SMILES strings, 2D graphs, or 3D coordinates. Such diverse representations are essential for capturing the intricate details of molecular interactions. However, the high-dimensional and heterogeneous nature of spectral data, combined with challenges such as noise, baseline drift, and instrument variability, demands robust preprocessing pipelines. Early work demonstrated that conventional normalization and alignment techniques were insufficient for fully preserving the chemical information embedded in these datasets. Recent studies, including those by~\citep{gastegger2017hdnnp} and~\citep{gerrard2020impression}, have underscored the importance of integrating domain-specific knowledge-such as physics-informed normalization and tailored feature extraction into the preprocessing stage. More recent work, such as~\citep{alberts2024leveraging,alberts2024unraveling}, is building a large-scale spectral dataset and harnessing the power of transformer-based models to map the latent representations of spectral data, thereby paving the way for robust and generalizable spectral analysis frameworks. These advances ensure that the learned representations are both resilient to experimental artifacts and chemically meaningful, thereby establishing a strong foundation for addressing both forward and inverse tasks. 

\subsection{Forward Problem: Molecule-to-Spectrum }

The forward problem in SpectraML aims at predicting spectral information directly from known molecular structures, serving as an efficient alternative to computationally expensive quantum-chemical simulations and laborious experimental measurements. 
Forward-problem can also be extended to extract critical spectral features and related chemical properties.
Therefore, the \textbf{input} of these ML-empowered solutions consists of molecules represented in different forms, such as SMILES strings, molecular graphs, or three-dimensional coordinates. The \textbf{output} can be either full spectra across different modalities (MS, NMR, IR, Raman, UV-Vis) or specific spectral features and chemical properties relevant to the target application. These 
\textbf{ML approaches} typically adopt %either an \textbf{encoding–decoding} scheme, which maps molecular structures to full spectra, or 
an \textbf{encoding–prediction} framework, which predicts spectral features or related chemical properties in forms of regression or classification.
In such architectures, the encoder transforms the molecular structure into a latent feature space that captures its essential chemical characteristics. The subsequent prediction stage then leverages this representation to predict partial spectra or specific spectral properties, depending on the target modality.
While encoding and prediction are often implemented and trained end-to-end within a single model, without a strict separation (as demonstrated in example tasks D and E in Fig. \ref{fig:task}), we structure the following discussion of related work based on the various forms of input and output involved in these problems, as they directly influence the selection and design of applicable machine learning models.

% Next, we discuss the encoder designs for handling different molecular representations, followed by the decoder/classifier architectures used for predicting various types of spectral outputs.

\textbf{Input  Encoding}. 
As summarized in Table \ref{tab:paper_summarization}, the input to the forward problem is often in the form of vector-based molecular features/descriptors, and 2D, or 3D molecular graphs. This determines the choice of encoding, which is typically implemented as an MLP for vector-based molecular features/descriptors and a GNN for 2D and 3D molecular graphs.
While vector representations are straightforward to handle, molecules represented as graphs require more sophisticated processing. Message-passing layers within GNNs effectively capture the structural and relational information between atoms and bonds~\citep{guan2021real,kwon2020neural,kang2020predictive,jonas2019rapid,guo2022graph}. These graph-based encoders are typically paired with regression or classification modules to predict continuous properties, such as $^1\mathrm{H}$ and $^{13}$C chemical shifts~\citep{guan2021real,kwon2020neural,kang2020predictive,jonas2019rapid}, or to learn spectral features like excitation energies and spectral line shapes~\citep{mcnaughton2023machine,chen2022uvvis,singh2022graph}. Alternatively, encoders may utilize direct coordinate-based features. For example, physics-informed neural networks extract vibrational properties directly from atomic coordinates--integrating experimental observations with quantum chemical insights--to predict key quantities such as dipole moment derivatives and polarizability tensors~\citep{schutt2021equivariant,gastegger2021machine,chen2024accelerating,sowa2024nnp}.

\textbf{Output Prediction}. 
The ``Task Type'' and  ``Output'' column in Table \ref{tab:paper_summarization} for the forward problem indicate that the output prediction is mostly in the form of regression. For example in MS prediction, molecular substructures are mapped to fragment \textit{m/z} values and intensities. In NMR spectroscopy, three-dimensional molecular graphs serve as inputs to predict continuous $^1\mathrm{H}$ and $^{13}$C chemical shifts~\citep{guan2021real,kwon2020neural,kang2020predictive,jonas2019rapid}, which in turn enables the accurate prediction of coupling constants and supports MAS-based spectral reconstructions~\citep{cordova2023pure}. For IR, Raman, and UV spectroscopy, the prediction stage typically employs regression or classification layers to generate vibrational properties. In particular, key quantities--such as dipole moment derivatives and polarizability tensors--are predicted to capture the essential physical characteristics of the molecules~\citep{schutt2021equivariant,gastegger2021machine,chen2024accelerating,sowa2024nnp}. For ultraviolet and electronic spectra,  excitation energies and spectral line shapes are predicted~\citep{mcnaughton2023machine,chen2022uvvis,singh2022graph}.
%The prediction stage translates the latent molecular representation into the desired spectral output.  In an \textbf{encoding–decoding} framework, the decoder generates full spectra--either as a continuous profile or as a sequence of spectral tokens~\citep{wei2019rapid}.  In an \textbf{encoding-prediction} framework, classification components are integrated into 
The prediction could also involve validating subformulas and predicting discrete spectral features~\citep{goldman2023prefix,zhu2020using,young2021massformer,park2024mass,zhu2023rapid,murphy2023efficiently,goldman2024generating}, with some approaches further extending the framework to perform joint property prediction~\citep{voronov2022multi}. The output can also take the form of a sequence of spectral tokens~\citep{wei2019rapid}. In this case, a generative model is employed to map a SMILES string to the full spectrum, enabling sequence-based spectral prediction. 

\textbf{Extension of Forward Problems}. 
%Forward-problem can be extended to extract critical spectral features and related chemical properties. 
Beyond simulating spectral profiles, these forward-modeling approaches also support property-focused tasks: classification models can reveal reaction behaviors in MS-based reactivity analyses~\citep{fine2020graph}, while hybrid ML-first-principles methods utilize IR data to infer adsorption energies and bond strengths~\citep{du2023spectroscopy}. Additional efforts have extended these frameworks to predict other physicochemical parameters, such as logD values at various pH levels~\citep{leniak2024nmr}, thereby supporting broader applications in drug discovery, catalyst design, and materials optimization.

\subsection{Inverse Problem: Spectrum-to-Molecule}

The inverse problem in spectral analysis aims at inferring a molecular structure directly from its measured spectrum, providing a complementary approach to traditional structure elucidation methods. In this task, the \textbf{input} consists of spectral measurements that can vary widely, from one-dimensional NMR signals and high-dimensional spectral vectors to image-like two-dimensional matrices (e.g., from NMR or IR) and sequential data from mass spectrometry (MS). The \textbf{output} is the predicted molecular structure, commonly represented as a molecular graph or an SMILES string. \textbf{ML approaches} to the inverse problem typically adopt either an \textbf{encoding–decoding} scheme, where the spectral data is transformed into a latent representation and then decoded into a molecular structure (e.g., task B  in Fig. \ref{fig:task}), or an \textbf{encoding–prediction} framework, which directly predicts molecular substructures or functional groups from the spectral features (e.g., task  C in Fig. \ref{fig:task}). In such architectures, the encoder processes the input spectra to capture the critical information necessary for structure elucidation, and the subsequent decoder or classifier reconstructs the corresponding molecular representation.

\textbf{Input Encoding}.
As presented in the bottom part of Table \ref{tab:paper_summarization}, the input to inverse modeling typically consists of  one-dimensional $^1$H or $^{13}$C NMR spectra, which are often represented as high-dimensional vectors. For example, \citep{hu2024accurate} employs a multitask, transformer-based model to encode 1D NMR signals into a latent space, facilitating the reconstruction of full molecular structures and substructure arrays. Similarly, \citep{huang2021framework} integrates convolutional neural networks with beam search to process spectral inputs, predicting substructure probabilities and iteratively assembling complete molecular graphs. In another approach, \citep{yao2023conditional} leverages a bidirectional, auto-regressive transformer (BART) ~\citep{lewis2019bart} that is pre-trained on large-scale molecular data and fine-tuned with $^{13}$C NMR constraints. Additional methods, such as that of \citep{alberts2023learning}, tokenize NMR spectral features into sequences for encoding, while \citep{sridharan2022deep} combines Monte Carlo Tree Search with graph convolutional networks to iteratively build molecular graphs guided by spectral cues. These encoder designs are crucial for capturing both local and global spectral patterns that underpin accurate molecular reconstruction.

\textbf{Output Decoding and Prediction.}  
When the output of the inverse problem is a molecular structure, the task becomes a \textbf{generative} problem, where the decoder functions as a generator to reconstruct the molecular structure from spectral data, either as a sequence of tokens representing SMILES strings or by progressively constructing molecular graphs. For instance, \citep{alberts2023learning} utilizes a transformer decoder to convert tokenized NMR or IR spectra into SMILES strings, treating each spectral absorption value as a sequence element in a translation-like process. Similarly, \citep{jonas2019deep} frames the molecule reconstruction task as a Markov decision process (MDP) and incrementally reconstructs the molecule through a relation network. Moreover, additional constraints can be incorporated to refine the generative process; for example, \citep{sun2024cross} couples generative models with contrastive retrieval to enhance candidate matching accuracy, \citep{zheng2024cbmaff} focuses on classifying seized substances from $^1$H and $^{13}$C NMR data, and \citep{tian2024enhancing} verifies proposed structures through joint analysis of image-like spectral data and graph-based molecular features.

\begin{table*}[t!]
\centering
\small
\scalebox{0.89}{
\begin{tabular}{@{}c|ccccc@{}}
\toprule

\multirow{2}{*}{\textbf{Paper}} & \multicolumn{5}{c}{\cellcolor{lightblue!30!white} \textbf{Forward Problem: Molecule-to-Spectrum Prediction}} \\ 
\cmidrule(lr){2-6}  
 & \textbf{Task Type} & \textbf{Input} & \textbf{Output} & \textbf{Model} & \textbf{Dataset} \\  
\midrule

\multicolumn{6}{c}{\textbf{Vector-Based Molecular Representations}} \\ 
\midrule
\citep{binev2007prediction} & REG & Molecular features & Chem. shift, Coupling const. & ASNN & Custom   \\ 
\citep{gastegger2017hdnnp} & REG & 3D coordinates & Simulated IR spectrum & MLP & Custom  \\ 
\citep{gerrard2020impression} & REG & Coulomb matrix & Chem. shift, Coupling const. & KRR & CSD subset  \\ 
\citep{ye2020proteinir} & REG & Coulomb matrix & IR properties & MLP & Custom  \\ 
\citep{chen2022uvvis} & REG & Bispectrum components & Vertical excitation energy & LASSO & Custom  \\ 
\citep{lin2022iontransport} & REG & Geometric descriptors & Chemical shift & MLP & Custom  \\ 
\citep{sowa2024nnp} & REG & Geometric descriptors & Polarizability tensor & KRR & Custom \\ 
\citep{homanh2024prediction} & REG & Molecular features & Vacuum UV spectrum & Random Forest & Custom \\ 
\midrule

\multicolumn{6}{c}{\textbf{2D Graph-Based Molecular Representations}} \\ 
\midrule
\citep{jonas2019rapid} & REG & 2D graph & Chemical shift & GNN & NMRshiftdb2 \\ 
\citep{kwon2020neural} & REG & 2D graph & Chemical shift & GNN & NMRshiftdb2  \\ 
\citep{kang2020predictive} & REG & 2D graph & Chemical shift & GNN & NMRshiftdb2  \\ 
\citep{zhu2020using} & REG & 2D graph & MS peaks vector & GNN & NIST17  \\ 
\citep{young2021massformer} & REG & 2D graph & MS peaks vector & Transformer & NIST20  \\ 
\citep{goldman2023prefix} & CLS, REG & 2D graph & Subformula classification & GNN & NIST20, NPLIB1 \\  
\citep{zhu2023rapid} & REG & 2D graph & MS peaks vector & GNN & NIST17  \\ 
\citep{murphy2023efficiently} & CLS, REG & 2D graph & Subformula classification & GNN & NIST20 \\
\citep{park2024mass} & REG & 2D graph & MS peaks vector & GNN & NIST20 \\ 
\citep{goldman2024generating} & REG & 2D graph & MS peaks vector & GNN & NIST20  \\ 
\midrule

\multicolumn{6}{c}{\textbf{3D Molecular Representations}} \\ 
\midrule
\citep{gastegger2021machine} & REG & 3D graph & Multiple spectral properties & GNN & MD17 \\ 
\citep{schutt2021equivariant} & CLS & 3D graph & Peptide-spectrum matches & GNN & MD17, QM9  \\ 
\citep{guan2021real} & REG & 3D graph & Chemical shift & GNN & NMRshiftdb2  \\ 
\citep{singh2022graph} & REG & 3D graph & Excitation spectrum & GNN & QM9  \\ 
\citep{chen2024accelerating} & REG & 3D graph & Energy, forces, dipole moments & GNN & Custom  \\ 
\midrule

\multicolumn{6}{c}{\textbf{SMILES Representations}} \\ 
\midrule
\citep{wei2019rapid} & GEN & SMILES Seq & EI-MS prediction & MLP & NIST17 \\ 
\midrule

\multirow{2}{*}{\textbf{Paper}}  & \multicolumn{5}{c}{\cellcolor{lightblue!30!white} \textbf{Inverse Problem: Spectrum-to-Molecule Prediction}} \\ 
\cmidrule(lr){2-6}
 & \textbf{Task Type} & \textbf{Input} & \textbf{Output} & \textbf{Model} & \textbf{Dataset}  \\ 
\midrule

\multicolumn{6}{c}{\textbf{NMR Spectral Representations}} \\ 
\midrule
\citep{jonas2019deep} & GEN & Formula + NMR vector & Molecule Graph & GNN & NMRshiftdb \\
\citep{sridharan2022deep} & GEN & NMR Vector & Molecule Graph & GCN & NMRshiftdb2 \\
\citep{yilmaz2022novo} & GEN & MS Seq & SMILES Seq & Transformer & DeepNovo \\
\citep{kim2023deepsat} & CLS & NMR image & Molecule structure classification & CNN & Custom \\ 
\citep{yao2023conditional} & GEN & NMR Seq & SMILES Seq & Transformer & CReSS \\
\citep{alberts2023learning} & GEN & NMR sequence & SMILES sequence & Transformer & Pistachio \\
\citep{hu2024accurate} & GEN & NMR vector  & SMILES sequence & Transformer & SpectraBase  \\  
\citep{leniak2024nmr} & REG & NMR vector & LogD value & SVR & SpecFAI   \\ 
\citep{yan2024resolution} & GEN & Low-resolution NMR image & High-resolution image & GAN & Custom \\ 
\citep{guo2024can} & GEN,REA & IR, NMR, MS image & SMILES sequence & MLLM & MolPuzzle  \\ 
\citep{su2024automation} & GEN,REA & BitMap image & Molecule Graph  & LLM & Custom \\
\midrule

\multicolumn{6}{c}{\textbf{Other Spectral Representation}} \\ 
\midrule
\citep{wei2019rapid} & REG & MS vector & Intensity values & MLP & NIST2017  \\ 
\citep{fine2020spectral} & CLS & IR/MS vector & Functional group classification & MLP & CANDIY  \\ 
\citep{fine2020graph} & CLS & MS vector & Reaction classification & Decision Tree & MoP  \\ 
\citep{enders2021functional} & CLS & IR image & Functional group classification & CNN & FTIRML \\ 
\citep{alberts2024leveraging} & GEN, REA & IR Seq & SMILES Seq & Transformer & NIST2010 \\ 
\bottomrule
\end{tabular}}
\caption{Summary of ML approaches in spectral analysis categorized into \textbf{Forward Problems} (molecule-to-spectrum) and \textbf{Inverse Problems} (spectrum-to-molecule). Studies are grouped by input representation. Task types are annotated as: CLS (Classification), REG (Regression), GEN (Generation), and REA (Reasoning).}
\label{tab:paper_summarization}
\vspace{-0.3in}
\end{table*}
Alternatively, an \textbf{encoding-prediction} framework maps the spectral representation to discrete structural elements, such as molecular substructures or functional groups, without generating an entire molecular structure. In this paradigm, the model deduces how atoms and functional groups are arranged to produce the observed spectral features, a capability that is particularly valuable for applications ranging from natural product identification to forensic analysis. For example, in MS context,  MEDUSA~\citep{boiko2022fully} and CANOPUS~\citep{duhrkop2020systematic} incorporate classification and ranking layers to discriminate between candidate metabolites based on MS and MS/MS features, while CandyCrunch~\citep{urban2024predicting} predicts glycan topologies by analyzing tandem MS data. In  IR spectroscopy,   CANDIY-spectrum~\citep{fine2020spectral} and CNN-based methods~\citep{enders2021functional} focus on identifying diagnostic functional groups from characteristic absorption patterns. Similarly, transformer-based networks for  NMR leverage $^1$H and $^{13}$C spectra to predict both key substructures and complete molecular formulas for robust classification and reconstruction~\citep{huang2021framework,hu2024accurate,alberts2023learning}. By directly extracting these critical features, the encoding-prediction paradigm offers an interpretable and efficient alternative to generative approaches for structure elucidation.

\textbf{Extension of Inverse Problems}.  
Beyond full structure elucidation, inverse SpectraML can be extended to recover detailed substructural information and functional group classifications that are essential for rapid compound identification and downstream analysis. For instance, sequence-to-sequence models applied to MS/MS data--such as those demonstrated in Casanovo~\citep{yilmaz2022novo}--and hybrid systems that combine substructure detection with full structure generation~\citep{kim2023deepsat,yao2023conditional} further enhance compound identification when integrated with spectral databases, as seen in CFLS~\citep{sun2024cross}. These extended approaches broaden the applicability of inverse SpectraML to diverse fields, from natural product discovery and metabolite screening to forensic investigations, thereby significantly reducing the reliance on time-intensive manual verification.

\subsection{Unified Frameworks and Cross-Modal Integration}

Recognizing that forward and inverse problems share common underlying chemical principles, recent research has begun to develop unified frameworks capable of addressing both tasks concurrently. \textbf{Foundation models} pre-trained on large, heterogeneous spectral datasets are at the forefront of this endeavor. These models leverage cross-domain learning to capture shared features across diverse modalities, such as IR, NMR, MS, and Raman--thereby enabling few-shot and zero-shot learning capabilities~\citep{bommasani2021opportunities}. Concurrently, physics-informed generative models, including diffusion models and GAN-based super-resolution techniques, have been introduced to synthesize high-fidelity spectra while respecting known chemical constraints~\citep{cordova2023pure}. Hybrid architectures that combine the relational modeling strength of GNNs with the sequence modeling capabilities of transformers offer a particularly promising route toward integrated spectrum analysis~\citep{young2021massformer}. Moreover, foundation models are steering the \textbf{advancement of reasoning-driven spectrum analysis}, particularly in complex inference tasks such as spectral deconvolution, peak assignment, and spectral consistency validation~\citep{guo2024can,su2024automation,alberts2024unraveling}. These models can reason about ambiguous spectra by leveraging prior chemical knowledge to infer plausible molecular structures, resolve overlapping spectral features, and predict missing spectral regions. By unifying forward and inverse tasks within a single framework, these emerging approaches not only alleviate issues of data scarcity through synthetic data generation \citep{tan-etal-2024-large, chen2024interleaved} but also enhance model robustness and interpretability. This integrated perspective is poised to accelerate discovery in diverse domains such as drug development, materials science, and environmental monitoring.

\section{Challenges and Opportunities}
\label{sec:discussion}

% The intersection of spectroscopy and machine learning presents both significant challenges and promising opportunities that are critical for advancing the field of SpectraML. In the following sections, we first discuss the primary challenges related to data acquisition, integration, and computational demands, and then outline emerging opportunities--including advanced integration techniques, synthetic data generation, and the transformative potential of foundation models.
%\vspace{-0.05in}
\subsection{Data Quality, Scarcity, and Complexity}

SpectraML faces several interrelated challenges arising from the inherent nature of experimental spectral data and the limitations of current ML approaches. First, the variability in data quality is a significant obstacle. Experimental spectra are often compromised by noise, baseline drifts, and instrument-to-instrument discrepancies, leading to inconsistencies in spectral resolution and intensity. Such variability complicates model training and can severely degrade the predictive performance of ML algorithms--especially when preprocessing pipelines are insufficiently robust. Moreover, the scarcity and imbalance of high-quality, annotated spectral datasets, particularly for rare or complex compounds, further exacerbate the issue. The limited availability of training data not only hinders the generalization of models across diverse chemical spaces but also increases the risk of overfitting, necessitating strategies such as data augmentation and transfer learning.

In addition to data quality challenges, the intrinsic complexity of spectral data presents a formidable hurdle. Spectral measurements typically exhibit high dimensionality and overlapping peaks, making feature extraction a nontrivial task. Current ML models often struggle to capture the nuanced, high-dimensional patterns inherent in such data, which leads to suboptimal performance in tasks like peak detection and feature discrimination. Furthermore, many existing architectures are not designed to fully leverage the domain knowledge embedded in spectroscopic data, thereby limiting their ability to exploit underlying chemical and physical principles.

A further challenge arises from the need to integrate data from multiple spectroscopic techniques (e.g., IR, MS, and NMR), each characterized by distinct scales, formats, and noise properties. Developing effective fusion strategies to reconcile these differences into a unified model is nontrivial. Most current ML architectures are optimized for single-modality inputs and often fail to capture the critical cross-domain relationships needed for accurate spectral analysis. Additionally, achieving model interpretability--so as to provide meaningful insights into the underlying chemical phenomena--requires a careful balance between model complexity and transparency. Addressing these challenges is crucial for advancing the state of SpectraML and ensuring that ML-driven approaches can fully harness the rich information contained in spectral data.

\subsection{Opportunities %in Advanced Data Integration 
and Emerging Paradigms}

%\subsubsection{Synthetic Data Generation and Physics-Informed Methods}

\textbf{Synthetic Data Generation and Physics-Informed Methods}.
One promising avenue to overcome the challenges posed by scarce and variable spectral data is advanced data augmentation using AI-driven generative models \citep{wu2024unigen}. State-of-the-art approaches, like large language models (LLMs) and diffusion models, enable effective application in many downstream tasks \citep{chen2025unveiling, guo2023can, chen2024uncertainty, chen2024gui, huang2023metatool}. They can learn complex, high-dimensional data distributions from experimental spectra \citep{luo2023fast,bushuiev2024massspecgym}. Once trained, these models can rapidly generate synthetic spectra that replicate the overall shape and key features of real data, while also capturing subtle nonlinear relationships that traditional simulation methods may overlook \citep{alberts2024leveraging,alberts2024unraveling}. This capability is particularly valuable in SpectraML, where limited annotated datasets and the high cost of experimental measurements hinder robust model training.

Moreover, incorporating physics-based priors into these generative frameworks can significantly enhance the chemical validity and interpretability of the synthetic spectra. By embedding domain-specific constraints--such as conservation laws, known peak intensity ratios, and chemical shift rules--into the generative process, the models are guided to produce outputs that adhere to established physical and chemical principles. This hybrid approach not only mitigates data scarcity by rapidly generating high-fidelity synthetic data but also offers computational advantages over conventional methods like density functional theory (DFT) or molecular dynamics simulations. Consequently, these advancements have the potential to accelerate discovery and innovation in fields ranging from drug development to materials science.

\textbf{Foundation Models: A New Paradigm for SpectralML}.
Foundationmodels~\citep{bommasani2021opportunities,moor2023foundation} represent a transformative opportunity in spectral analysis by being pre-trained on extensive, heterogeneous datasets that span multiple modalities (e.g., IR, NMR, MS, and Raman). These models leverage cross-domain learning to capture both global chemical phenomena and local spectral details, enabling few-shot and zero-shot learning to rapidly adapt to tasks such as spectral generation, peak prediction, and structure elucidation. Their ability to handle both forward and inverse tasks, coupled with the integration of domain-specific priors like conservation laws and chemical shift rules, paves the way for unified analytical frameworks built on robust pre-training data and carefully tailored model architectures.

The advanced reasoning capabilities inherent in foundation models further enhance spectral analysis by enabling multi-step inference and hypothesis generation that effectively address ambiguous or overlapping spectral features. By integrating information across multiple spectral modalities, the models can construct a holistic view of molecular structures while dynamically adjusting predictions in response to new data. This adaptive process, combined with built-in error detection and uncertainty quantification, not only ensures chemically plausible outcomes but also accelerates discovery through iterative learning and the simulation of "what-if" scenarios. Together, these features foster a more explainable and efficient approach to both traditional and emerging spectral tasks. Moreover, when utilizing these models, it is also essential to consider potential trustworthiness issues \citep{huang2024trustllm, huang2023trustgpt}, including hallucination \citep{ji2023survey, gao2024honestllm}, inconsistency \citep{huang2024social}, safety \citep{zhou2024labsafety, huang2024obscureprompt}, and robustness \citep{huang2025breaking}, as these factors directly impact their reliability and ethical alignment in real-world applications.

% \subsubsection{Automation, Interpretability, and Hybrid Modeling}
% In addition to improved data integration, emerging ML approaches are driving automation and enhancing interpretability in spectral analysis. By incorporating explainable AI techniques, advanced models can provide interpretable insights into the chemical phenomena underlying spectral data. This capability enables the automation of complex tasks such as peak assignment, spectral deconvolution, and molecular structure prediction, reducing reliance on manual expert interpretation and supporting real-time decision-making in experimental workflows. Moreover, hybrid modeling approaches that combine traditional ML methods--known for their computational efficiency and transparency--with deep learning architectures promise to balance predictive accuracy with interpretability. Advances in transfer learning and model compression further reduce computational overhead, making sophisticated SpectraML methods more accessible to a broader research community.

\section{Conclusion}
\label{sec:conclusion}

The SpectraML establishes a crucial intersection between machine learning and spectroscopy. In this work, we provide a comprehensive overview of SpectraML and present a unified roadmap that traces methodologies across multiple spectroscopic techniques and categorizes key advancements in forward and inverse problems. To support future research, we highlight emerging trends such as generative modeling and foundation models and release an open-source repository. This survey serves as a valuable resource for researchers in both chemistry and AI, fostering interdisciplinary collaboration and driving innovation in spectral analysis.

\section*{Acknowledgments}

This work was supported by the National Science Foundation under the NSF Center for Computer Assisted Synthesis \href{https://ccas.nd.edu/}{(C-CAS)}, grant number CHE-2202693.
\newpage
\bibliography{icml2024}
\bibliographystyle{icml2024}

\newpage
\section{Appendix}
\label{sec:Spectroscopy-data}

This section provides a quick summary of the different spectra mentioned in this survey, focusing on defining the spectra.

\subsection{Mass Spectrometry (MS)}

Mass Spectrometry (MS) is used for the determination of the molecular mass and molecular formula of a compound and substructures (fragments) produced when the molecule breaks apart that provide information on some of its structural features. The spectrum is usually presented as a vertical bar graph, in which each bar on the x-axis represents an ion that has a specific mass-to-charge ratio (\textit{m/z}) and the length of the bar indicates the relative abundance of the ion on the y-axis. The peak with the highest \textit{m/z} value is called the molecular ion (M), which gives the molecular mass of a compound. The peaks with smaller \textit{m/z} values are called fragment ion peaks, which are the charged fragments of the molecular ion (M). The tallest peak is called the base peak because it has the greatest relative abundance as a stable charged fragment. The base peak is assigned a relative abundance of 100\%, and the relative abundance of each of the other peaks is shown as a percentage of the base peak. The mass spectrometer records all the relative abundance of each fragments plotted against its \textit{m/z} value into an MS spectrum. 

\subsection{Infrared (IR) and Raman spectrum}
An IR  and the related Raman spectra probe the vibrations along bonds, angles and torsions in a molecule through their interactions with electromagnetic radiation in the range of ~4000–~600 cm\textsuperscript{-1}. The results are presented as a line graph of the percent transmission of radiation on the y-axis and the wavenumber (or wavelength) of the radiation transmitted on the x-axis. At 100\% transmission all the energy of the radiation of a particular wavelength passes through the molecule. Lower values of percentage transmission mean that some of the energy is being absorbed by the compound. Each downward spikes (called bands or peaks) in the spectrum represents the absorption of energy. A Fourier transform IR (FT-IR) spectrophotometer measures all frequencies simultaneously. This allows for multiple measurements of the sample, which are then averaged. The information is then digitized and Fourier transformed by a computer to produce the final FT-IR spectrum. Within the IR/Raman spectrum, there are two major distinguishable areas. The area on the left (4000–1400 cm\textsuperscript{-1}) is where most functional groups show absorption bands, called the functional group region because these vibrations are highly localized and specific for certain functional groups. The area on the right (1400–600 cm\textsuperscript{-1}) is called the fingerprint region because it is characteristic of the compound as a whole. The main IR analysis is on the functional group region, while the fingerprint region is hard to interpret but can be used to match the structure of a compound to a known reference spectrum. 

\subsection{UV-Vis spectrum}
A UV-Vis spectrum measures electronic transition in a molecule, a molecule absorbs either ultraviolet (180-400 nm wavelength) or visible light (400-780 nm wavelength). It is presented as a graph of absorbance, optical density or transmittance on the y-axis as a function of wavelength on the x-axis. This spectroscopy provides information about compounds that have conjugated double bonds, because double bonds have molecular orbitals that can potentially absorb the applied energy. 

\subsection{Nuclear Magnetic Resonance (NMR) spectra}

NMR data provides information on certain nuclei, including the hydrogen and carbon atoms that make up most organic molecules. To record an NMR spectrum, a sammple is introduced in a constant, strong magnetic field and an electromagnetic pulse of short duration covers a range of frequencies in the range of 60-900 MHz. Each nucleus of a compound sample can absorb the frequency it requires to come into resonance and produce a signal called a free induction decay (FID) at a frequency corresponding to the change in energy. The intensity of the FID signal decays as the nuclei loose energy they gained from an rf pulse. A computer then measures the change in intensity over time and converts it using a mathematical operation known as Fourier transform to produce a Fourier transform NMR (FT–NMR) spectrum. This is the final NMR spectrum that is then observed and analyzed by experimentalists and reported in publications and databases. The advantage of NMR spectroscopy over the other instrumental techniques is that it not only identifies the functionality at a specific carbon, but it also enables to obtain information about the connectivity of neighboring carbons. Hence, NMR spectra are much more diverse and complex. First, the spectrum can be in multiple dimensions, indicating the interaction between multiple (1D, 2D, 3D, or 4D-NMR). Secondly, NMR can represent different atomic nuclei. Some elements have a property called  nuclear spin that allows them to be studied by NMR such as hydrogen, carbon, nitrogen, fluorine, and phosphorous (\textsuperscript{1}H, \textsuperscript{13}C, \textsuperscript{15}N, \textsuperscript{19}F, \textsuperscript{31}P-NMR, respectively). Thirdly, 1D-NMR data is typically represented as a spectra plot where individual signals appear as peaks, with the x-axis showing the chemical shift (ppm), which is a measure of how a nucleus is shielded, and the y-axis representing the signal intensity. Furthermore, the area under each peak is proportional to the number of equivalent nuclei contributing to that signal called integration, and the shape of a peak can be complex due to coupling with neighboring nuclei, resulting in a splitting pattern of the peaks known as a multiplicity (either as a singlet, doublet, triplet, etc.) that contains information on the adjacent atoms.  
\end{document}